\documentclass[conference]{IEEEtran}
\IEEEoverridecommandlockouts
\usepackage{cite}
\usepackage{amsmath,amssymb,amsfonts}
\usepackage{algorithmic}
\usepackage{graphicx}
\usepackage{textcomp}
\usepackage{xcolor}
\usepackage{hyperref,titlesec,subcaption,diagbox,makecell,indentfirst}

\IEEEsettextwidth{0.75in}{0.75in}
\IEEEsetsidemargin{i}{0.75in}
\IEEEsettextheight{0.78in}{0.75in}
\def\IEEEtitletopspace{0.27in}
\IEEEsettopmargin{t}{0.78in}

\newsavebox\mybox
\newsavebox\myboxtwo

\def\BibTeX{{\rm B\kern-.05em{\sc i\kern-.025em b}\kern-.08em
    T\kern-.1667em\lower.7ex\hbox{E}\kern-.125emX}}
\begin{document}

\title{BlanketGen - A synthetic blanket occlusion augmentation pipeline for MoCap datasets
\thanks{This work is financed by National Funds through the Portuguese funding agency, FCT - Fundação para a Ciência e a Tecnologia, within project LA/P/0063/2020 as well as under the scope of the CMU Portugal (Ref PRT/BD/152202/2021).}
}


\setlength{\parskip}{0pt}
\titlespacing*{\subsubsection}
{0pt}{.4ex plus 0ex minus .2ex}{0ex plus 0ex}
\titlespacing*{\subsection}
{0pt}{.6ex plus 0ex minus .3ex}{0ex plus 0ex}
\titlespacing*{\section}
{0pt}{.8ex plus 0ex minus .4ex}{0ex plus 0ex}

\author{\IEEEauthorblockN{João Carmona\IEEEauthorrefmark{1}\IEEEauthorrefmark{2}, Tamás Karácsony\IEEEauthorrefmark{1}\IEEEauthorrefmark{2}\IEEEauthorrefmark{3}, João Paulo Silva Cunha\IEEEauthorrefmark{1}\IEEEauthorrefmark{2} \IEEEmembership{Senior Member, IEEE}}
\IEEEauthorblockA{\IEEEauthorrefmark{1}Center for Biomedical Engineering Research, INESC TEC, Porto, Portugal}
\IEEEauthorblockA{\IEEEauthorrefmark{2}Faculty of Engineering (FEUP), University of Porto, Porto, Portugal}
\IEEEauthorblockA{\IEEEauthorrefmark{3}Robotics Institute, Carnegie Mellon University, Pittsburgh, PA, USA}
}

\maketitle

\begin{abstract}
Human motion analysis has seen drastic improvements recently, however, due to the lack of representative datasets, for clinical in-bed scenarios it is still lagging behind. To address this issue, we implemented BlanketGen, a pipeline that augments videos with synthetic blanket occlusions. With this pipeline, we generated an augmented version of the pose estimation dataset 3DPW called BlanketGen-3DPW. We then used this new dataset to fine-tune a Deep Learning model to improve its performance in these scenarios with promising results. Code and further information are available at \href{https://gitlab.inesctec.pt/brain-lab/brain-lab-public/blanket-gen-releases}{https://gitlab.inesctec.pt/brain-lab/brain-lab-public/blanket-gen-releases}.
\end{abstract}


\section{Introduction}
\label{sec:intro}

Human motion analysis is an active research area that has recently seen drastic advancements by making use of Deep Learning (DL), with which incredible results have been attained \cite{DLHPEreview}. Due to this, it has become a hot topic in other areas of research as a tool to help solve complex problems related to the human body.

One such area is the semiology of epileptic seizures: accurate and effective diagnosis and classification of epilepsy requires visiting an Epilepsy Monitoring Unit (EMU), where the patients are monitored with video-electroencephalogram (video-EEG) systems; the outputs of these systems during epileptic seizures are then subjectively analyzed by epileptologists. Our group has been exploring automatic human motion analysis to aid epileptologists with quantitative results \cite{karacsonyClassification2020, neurokinect, karacsonyNature}, however acquisition of the clinical data required for it is challenging \cite{deepepil}.

Human Pose Estimation (HPE) is a task within the broader concept of human motion analysis that focuses exclusively on the position of the subjects. Different approaches have been studied to tackle the task of HPE, but for the specific use-case of seizure semiology in EMUs video-based systems are the most promising solutions.

However, most of the research effort invested into video-based HPE has been focused on the most common scenarios of subjects standing up and moving with few occlusions \cite{hpereview2021}; whereas the scenario considered in this paper has the subjects lying down in beds, usually with blankets covering them at least partially, and recorded with a fixed camera. The blanket occlusions in particular are of concern since accurate estimation of occluded joints is especially difficult. Current state-of-the-art systems generally make use of DL which is extremely dependent on the datasets used for training and blanket occlusions are rarely present in the datasets used for training \cite{hpereview2021}.

In order to allow DL HPE systems to make use of the information hidden in blanket occlusions, we propose BlanketGen, a pipeline to augment a dataset with computer-generated (CG) blanket occlusions. We used this pipeline to generate BlanketGen-3DPW, a version of 3DPW \cite{3DPW} augmented with CG blanket occlusions. We then used this new dataset to fine-tune a DL HPE model to improve its performance in these scenarios with promising results.

\section{Related Works}
Older approaches to video-based HPE used traditional computer vision techniques, such as the system proposed in \cite{neurokinect} for clinical in-bed HPE, which uses optical flow to automatically track manually selected masks that surround the body parts of interest. 
However, more recently the focus of general HPE research has shifted to improving DL HPE systems. 

In \cite{smpl} a human body model was proposed which used principal component analysis to describe variations in body shape with as few parameters as possible, this approach proved extremely robust and efficient so the SMPL model became a standard in HPE; \cite{keepItSMPL} then used 2D joint positions acquired with the method proposed in \cite{deepCut} to optimize a SMPL mesh; in \cite{hybrik} inverse kinematics were used to fuse 3D joint positions and SMPL parameters that were estimated by separate systems; \cite{dynaboa} employed online unsupervised learning to adapt to data in new domains even without ground truth annotations; \cite{deciwatch} used DL to interpolate pose estimations between keyframes which allowed for state-of-the-art results even while only analyzing one-tenth of the frames. 
Due to these and other developments, DL HPE systems are now exceptional in scenarios where there are few occlusions. However, their performance drops drastically when blanket occlusions are present.

In order to improve DL systems in such scenarios, \cite{patientMoCap} experimented with different approaches to improve HPE systems in clinical scenarios using depth video, among them synthetic blanket occlusions; \cite{clever2021} augmented depth data with synthetic blanket occlusions to train a DL system that estimated body shape, pose, and pressure maps of people laying down in beds. Both showed marked improvements when synthetic blanket occlusion augmentations were used during training, however, both worked only with depth data.

Very recently, the SLP dataset \cite{slp} was introduced, which includes 14.7 thousand RGB, depth, and LWIR images as well as pressure mat data of people lying in beds in different resting positions with and without blanket occlusions. It is an excellent resource, but it is a relatively small dataset and does not have the temporal information present in video.

In \cite{blanketset} we introduced BlanketSet, an action recognition dataset of people moving under blankets in an EMU, and it was used to evaluate the fine-tuned model proposed in this paper by means of a public survey. It was found that the fine-tuned model slightly improved the performance when real full-body blanket occlusions were present.

\section{Methods}
For the implementation of BlanketGen, the 3D modeling tool Blender \cite{blender} was used because it is free, has robust cloth simulation, and provides a Python API which supports most of the capabilities of the program.

3DPW \cite{3DPW} was chosen as the dataset to augment since it has a reasonable size (\(>\)50K frames), provides ground truth body models, and there are state-of-the-art DL HPE systems with open source code that used it for training and testing.

\begin{figure}[htpb]
\centering
\includegraphics[width=0.95\linewidth]{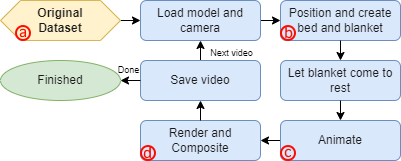} 
\caption{Diagram of the BlanketGen pipeline. The letters in red refer to the stages portrayed in Fig. \ref{fig:pipelineexample}.}
\label{fig:pipeline}
\end{figure}

BlanketGen processes the videos from 3DPW to generate new videos with CG blankets simulated on top of the subjects.
The pipeline is illustrated in figures \ref{fig:pipeline} and \ref{fig:pipelineexample}, and it is detailed in the following subsections.


\savebox{\mybox}{ \includegraphics[height=16em]{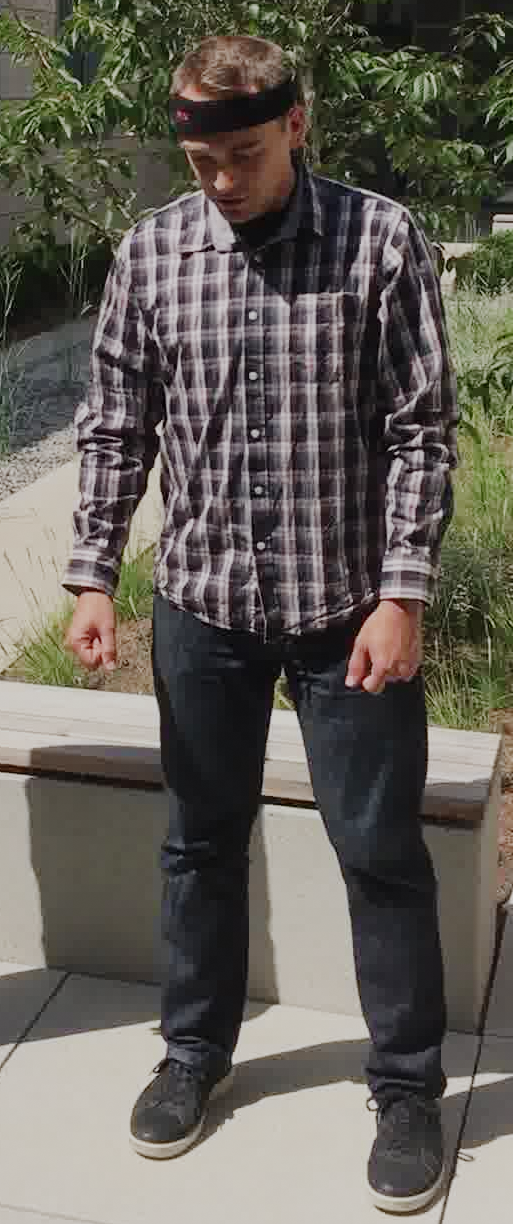}}
\savebox{\myboxtwo}{ \includegraphics[height=16em]{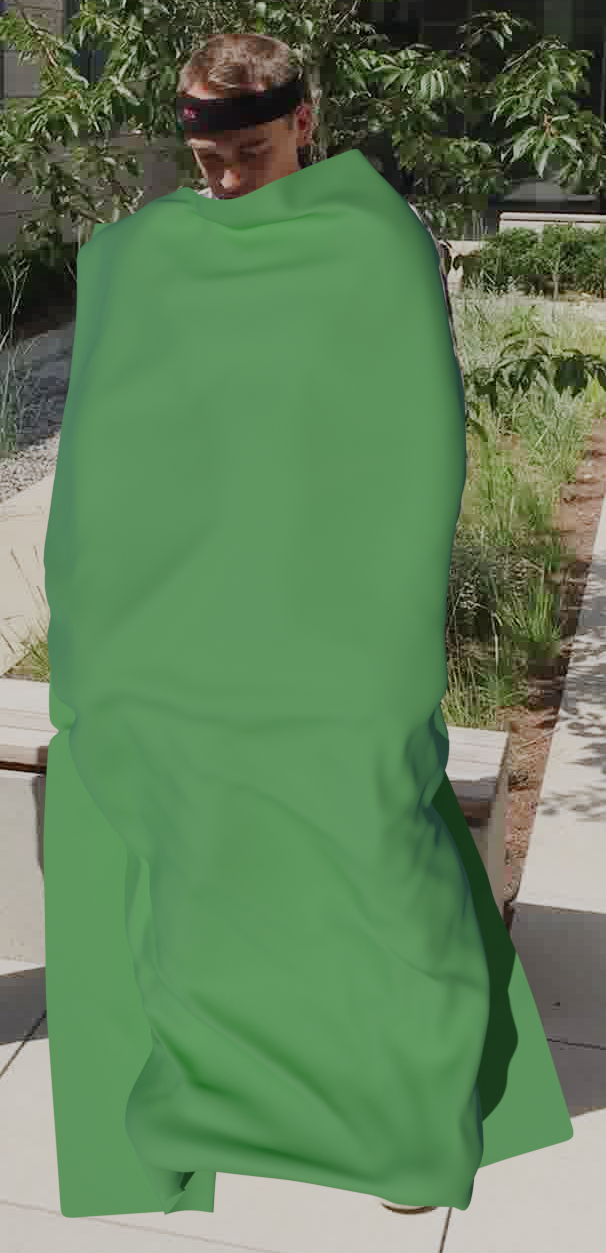}}

\begin{figure}[t]
\begin{subfigure}[t]{.49\linewidth}
  \centering
  \usebox{\mybox}
  \caption{Original frame from 3DPW}
  \label{fig:ogframe}
\end{subfigure}
\begin{subfigure}[t]{.49\linewidth}
  \centering
  \vbox to \ht\mybox{%
      \vfill
      \includegraphics[height=9.5em]{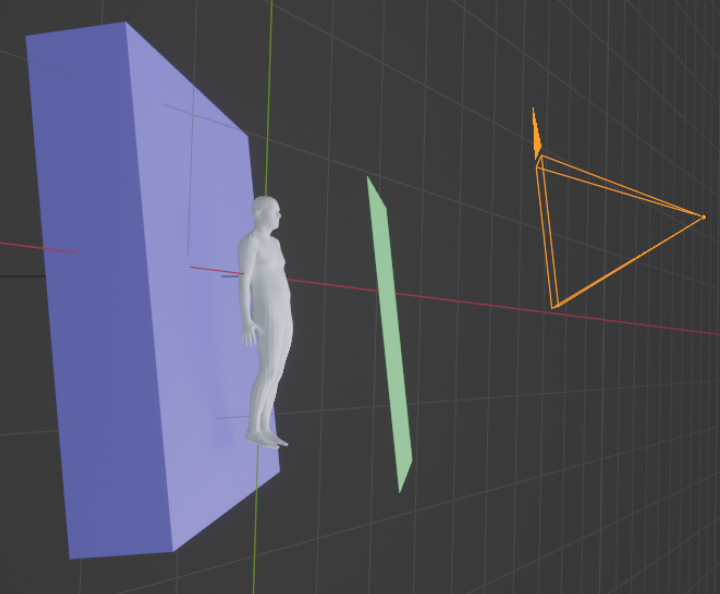}  
      \vfill
  }
  \caption{Scene after positioning}
  \label{fig:pframe}
\end{subfigure}
\begin{subfigure}[t]{.49\linewidth}
  \centering
  \vbox to \ht\myboxtwo{%
  \vfill
  \includegraphics[height=16em]{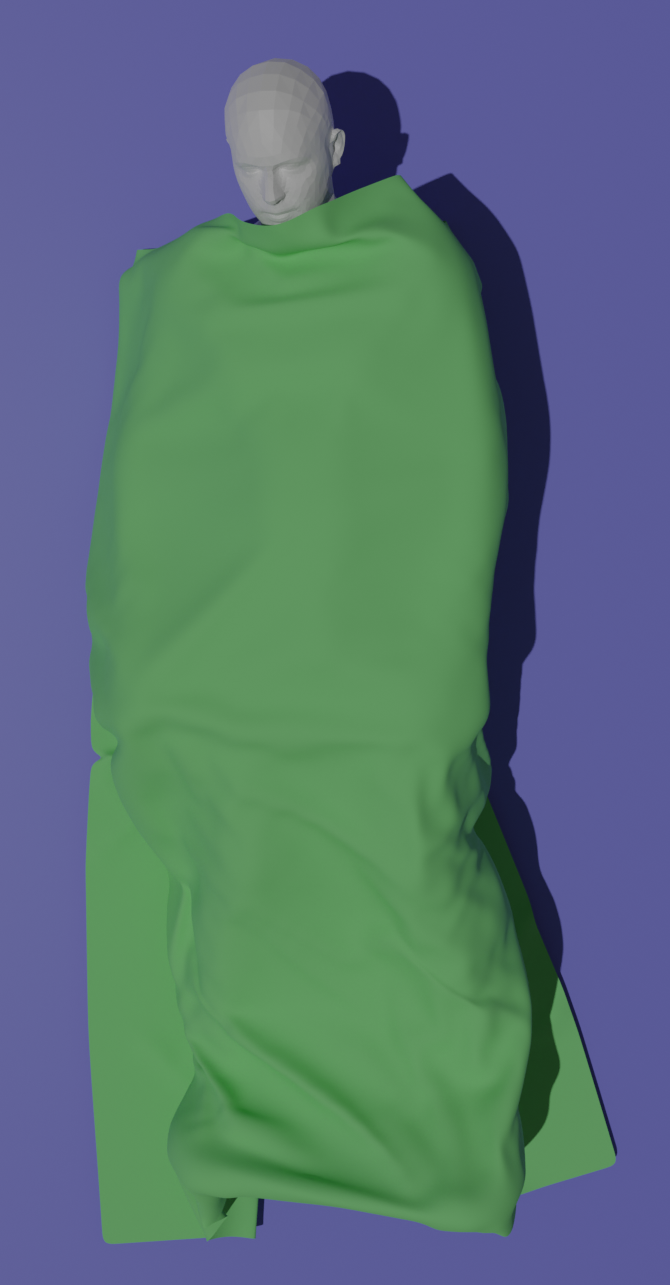}  
  \vfill
  }
  \caption{Simulated, rendered scene}
  \label{fig:rframe}
\end{subfigure}
\begin{subfigure}[t]{.49\linewidth}
  \centering
  \usebox{\myboxtwo} 
  \caption{Final composited frame}
  \label{fig:fframe}
\end{subfigure}
\setlength{\belowcaptionskip}{-10pt}
\caption{An example of different stages of the BlanketGen pipeline.}
\label{fig:pipelineexample}
\end{figure}

\subsection{Data Loading}

In order to load the SMPL models provided as ground truth onto Blender, BlanketGen uses the SMPL-X \cite{SMPL-X} python package to calculate the vertices and faces of the model.

To avoid collision errors caused by the SMPL models intersecting with each other, which would negatively affect the cloth simulation, only one model is loaded for each augmented video. This leads to the blanket only interacting with one of the subjects, the videos containing two subjects are generated separately for each subject, which is a practical way to avoid catastrophic cloth simulation errors.

A camera is created and added to the scene, its attributes are then altered to match the intrinsics matrix provided by 3DPW.
A sun light, which is a light source that shines parallel rays of light is also added.

\subsection{Positioning}
While the positions provided as ground truth in 3DPW are accurate relative to each other, the global positions have low-frequency errors and so the models drift over time.
To address this, the model of the subject is fixed at position (0,0,0) and the camera is re-positioned so that their relative positions remain correct.

The center of the body model in Blender does not match the one in the ground truth, so it has to be adjusted. This is done by iterating through all the vertices of the model and displacing them by the offset necessary for their location relative to the center of the model to be correct.

While iterating through all the vertices, the vertex most distant from the camera is found. A new set of axes is created with this vertex at the center and with one of the axes pointing directly away from the camera and another straight up from the camera's perspective. A rectangle cuboid bed is then placed in the scene using these axes, this way the distance between the bed and the body model can be a defined constant; this is done to avoid the subject's model clipping into the bed and causing the cloth simulation to fail.

The position of the simulated blanket is initially set a short distance away from the body model in the direction of the camera, and its orientation is set to match the orientation of the bed.

\subsection{Cloth simulation}
In order to create a realistic blanket, first, a rectangular plane is added, then this plane is divided into 5625 (\(75^2\)) smaller rectangles, the \textit{subdivision} modifier is also used, which further subdivides the plane but does not affect the cloth simulation. 

The blanket is then set as a cloth with the collision calculations being performed in 15 increments per frame, and the collision calculations being performed in 10 increments per frame. The collision distance is set to half a millimeter and the gravity is set perpendicular to the bed.

The cloth simulation is initialized by running the simulation for 24 frames before the video starts in order to let the blanket come to a rest on top of the body model.

The simulation of the blanket is updated frame by frame and at every frame the vertex closest to the body model is calculated, if the distance between this vertex and the body model is larger than a predefined threshold the simulation is stopped to prevent videos generated where the blanket falls off the subject. Then the simulation is re-initialized and the video generation continues. To avoid the pipeline getting stuck in cases where the blanket falls off either during the time the blanket is
initialized or where it glitches through the body model, the next video is started a minimum of 48 frames from the previous starting point.

\subsection{Rendering and Compositing}
Rendering is carried out with Blender's \textit{Cycles} engine, which uses raytracing to generate physics-based renders. Due to the ease of parallelization of Monte Carlo estimations, the render engine can use GPUs to speed up considerably \cite{rayGems}. A maximum of 1000 samples are calculated per pixel, with a time limit of 10 seconds per frame.

The material of the blanket uses the \textit{Principled BSDF} shader, which is the default option in Blender and is generally fairly robust. The color is randomized uniformly in the RGB space for each video generated.

The bed and the body model are set as holdouts, which makes them function as transparent masks. This means that when compositing the rendered scene with the original frames, only the blanket is visible, and when the blanket goes behind the body model from the camera's perspective the portion that's behind the body model becomes invisible.

Blender's node tree system was used to compose the scene, where an alpha-over node is used to layer the rendered scene on top of the original video.

\subsection{Format} 
The output of the compositor is saved as a sequence of JPG images in a folder for every video. Each video folder is contained within a folder for the original video from which it was generated. Furthermore, these folders are separated into three folders for the three dataset partitions (train, test, and validation).

\subsection{Evalutation}
HybrIK \cite{hybrik} was chosen to evaluate whether CG blanket occlusion augmentations could improve existing DL HPE systems. It was selected over other state-of-the-art alternatives as it has an open-source implementation with training code as well as a pre-trained model. 

The pre-trained model provided was originally trained on a joint dataset consisting of training data from Human3.6M \cite{h36m}, MPI-INF-3DHP \cite{3dhp}, 3DPW \cite{3DPW}, and MS COCO \cite{mscoco}. 

For the initial evaluation of BlanketGen it was then trained only on the training data from BlanketGen-3DPW for 10 epochs with a learning rate of $1\mathrm{e}{-3}$, then 10 epochs with a learning rate of $1\mathrm{e}{-4}$, and finally 10 epochs with a learning rate of $1\mathrm{e}{-5}$.

The performance of this fine-tuned model was compared with the original pre-trained model, both on 3DPW and BlanketGen-3DPW.

The impact of the CG blanket occlusion training on real-world blanket occluded data, was out of the scope of this paper, thus it is described in \cite{blanketset}, along with the dataset acquired for this purpose.

\section{Results}
\subsection{BlanketGen-3DPW}
BlanketGen-3DPW is an augmented dataset generated with the BlanketGen pipeline. It took 3 weeks to generate running on two desktop PCs using three Nvidia GPUs (two GTX 1080 TIs and one RTX 3060). In total it contains 1037 generated videos with synthetic blanket occlusions, a total of 96399 frames.

The pipeline generated a total of 1275 videos (134422 frames), but 238 (38023 frames) were manually excluded, 66 (14332 frames) due to failures in the cloth simulation, and 172 (23691 frames) due to inaccuracies caused by errors in the ground truth.

BlanketGen-3DPW was formatted to be used with HybrIK as a drop-in replacement for 3DPW. Therefore, it uses the same MS COCO format. 

\subsection{Evaluation}
Quantitative evaluation was done by testing both models on the testing sets of BlanketGen-3DPW and 3DPW. Some videos on 3DPW have two participants and only one is occluded in each generated video, so BlanketGen-3DPW has both occluded and unoccluded subjects. The unoccluded subjects were excluded from the evaluation, therefore the results on BlanketGen-3DPW include only occluded subjects.

The average PA-MPJPE was recorded for all test cases and is reported in table \ref{tab:quantitativetable}.

\begin{table}[h!]
    \centering
    \resizebox{0.45\textwidth}{!}{%
    \begin{tabular}{c|cc}
        \diagbox{Model}{Dataset} & \makecell{3DPW\(\downarrow\) \\ (mm)} & \makecell{BlanketGen-3DPW\(\downarrow\) \\ (mm)}\\

        \hline
        Pre-trained & \textbf{44.97} & 56.43\\
        Fine-tuned (ours) & 50.95 & \textbf{53.96} \\
    \end{tabular}}
    \caption{Procrustes Analysis Mean Per Joint Position Error (PA-MPJPE) in mm. Procrustes Analysis consists of a series of linear transformations to emphasize the shape rather than the rotation, location, or scale.}
    \label{tab:quantitativetable}
\end{table}

\section{Discussion}
There was an improvement in accuracy when blanket occlusions were present, however it came at the cost of a comparable decrease in accuracy when they weren't.

The results of both models on occluded subjects were still very good, this is probably due to the blankets in BlanketGen-3DPW mostly only occluding the legs and feet, and not the upper body. In \cite{blanketset} it was found that the fine-tuned model significantly outperformed the pre-trained one when full body occlusions were present, but not when only half the body was occluded; not only that, but the results when the occlusions covered the full body were much worse than with only partial occlusion. Those results paired with the ones in this paper suggest that including synthetic full-body occlusions in BlanketGen-3DPW might lead to a bigger improvement.

The decrease in accuracy on unoccluded subjects could be caused by the 30 epochs of fine-tuning not including data from the other datasets that the model was originally trained on, as well as the fairly low number of epochs of fine-tuning. 

BlanketGen-3DPW is a version of 3DPW with synthetic blanket occlusions and as such can fulfill the same purposes as 3DPW, but in contexts where blanket occlusions are present. This could prove useful in clinical settings where the patients are lying down in beds, such as epilepsy monitoring and sleep analysis.

\section{Conclusion}
Blanket occlusions reduce the performance of vision-based HPE systems whenever they are present; this is an issue for HPE in epilepsy monitoring units, where patients are often lying under blankets. 

The work presented in this paper explored the potential of augmenting HPE datasets with synthetic blanket occlusions to address this problem. A pipeline to augment the 3DPW dataset was implemented and an augmented dataset called BlanketGen-3DPW was generated; BlanketGen-3DPW was then used to fine-tune a DL HPE model that was originally trained on several datasets without blanket occlusions. This fine-tuned model and the original model were evaluated on the test set of BlanketGen-3DPW and the results were separated based on whether the subject was occluded or not.

The evaluation results were promising, although there is still room for further research. 
In \cite{patientMoCap} it was found that Recurrent Neural Networks had a particularly large performance boost from blanket occlusion augmentations and it was hypothesized that this was due to the temporal information in the blanket, therefore it is reasonable to expect DL HPE systems that make use of temporal information would see a bigger improvement. Improvements could also be attained by fine-tuning for longer with more datasets, as well as by including full body occlusions in BlanketGen-3DPW. It would also be worthwhile to compare these results with other forms of occlusion augmentation.

\bibliographystyle{IEEEtran}
\bibliography{main}


\end{document}